# Ensemble Framework for Cardiovascular Disease Prediction

Achyut Tiwari[1], Aryan Chugh[2], Aman Sharma[3]

Department of Computer Science & Engineering, Jaypee University of Information[1,2,3] Technology, Waknaghat, District Solan, Himachal Pradesh 173234, India

<u>achyuttiwari@acm.org</u>[1], <u>aryanchugh0509@gmail.com</u>[2], amans.3008@gmail.com[3]

Abstract

Heart disease is the major cause of non-communicable and silent death worldwide. Heart diseases or cardiovascular diseases are classified into four types: coronary heart disease, heart failure, congenital heart disease, and cardiomyopathy. It is vital to diagnose heart disease early and accurately in order to avoid further injury and save patients' lives. As a result, we need a system that can predict cardiovascular disease before it becomes a critical situation. Machine learning has piqued the interest of researchers in the field of medical sciences. For heart disease prediction, researchers implement a variety of machine learning methods and approaches. In this work, to the best of our knowledge, we have used the dataset from IEEE Data Port which is one of the online available largest datasets for cardiovascular diseases individuals. The dataset isa combination of Hungarian, Cleveland, Long Beach VA, Switzerland & Statlog datasets with important features such as Maximum Heart Rate Achieved, Serum Cholesterol, Chest Pain Type, Fasting blood sugar, and so on. To assess the efficacy and strength of the developed model, several performance measures are used, such as ROC, AUC curve, specificity, F1-score, sensitivity, MCC, and accuracy. In this study, we have proposed a framework with a stacked ensemble classifier using several machine learning algorithms including ExtraTrees Classifier, Random Forest, XGBoost, and so on. Our proposed framework attained an accuracy of 92.34% which is higher than the existing literature.

## 1. Introduction

Heart diseases remain the major cause of death in the world [1]. According to the World Health Organisation (WHO), In 2019 32% of total deaths were caused by cardiovascular diseases [2]. In India, the Ministry of Health and Family welfare reported cardiovascular diseases (CVD) contributed to 28.1% of total deaths [3-4]. India currently has the highest rate of the acute coronary syndrome and ST-elevation myocardial infarction (MI) in the world. Hypertensive heart disease is a significant issue in India, with 261,694 fatalities in 2013. (An increase of 138 per cent as compared to 1990) [3]. Blood tests, electrocardiograms, and other testings can help doctors identify the severity of a patient's cardiovascular illness (EKG or ECG), MRI (cardiac magnetic resonance imaging), and CT (cardiac computerized tomography). There may not be enough doctors accessible in poor nations to perform diagnostic tests on patients with cardiovascular disease, or the diagnostic tests may be inaccurate due to a lack of infrastructure, resulting in further difficulties and putting patients' lives in danger [5]. In India, there are only 91 doctors for every 1 lakh citizen [6]. The early detection of cardiovascular disease in patients, as well as the provision of appropriate treatment, reduces the number of premature deaths [7]. The ability to predict cardiovascular disease can have a significant impact on both the medical industry and people's lives. However, there are several forms of cardiovascular diseases starting with coronary heart disease. Coronary Heart Disease is a prevalent and well-known heart condition. This condition causes the coronary arteries to narrow or get damaged. Arteries provide the heart with essential nutrients and oxygen, but they are unable to do so properly owing to the formation of plaque containing cholesterol in the arteries. The second one is Heart Failure also known as congestive heart failure. In this situation, the heart is unable to pump blood properly to various parts of the body. This might be an advanced form of coronary artery disease, which causes the heart to become so weak that it is impossible to pump blood. Another is Congenital Heart Disease; this sort of cardiac disease is present from birth. As there can be holes between the two halves of the heart, also known as septal defects. Cyanotic heart disease is characterized by obstructive defects, which indicate that blood flow is totally or partially impeded via various parts of the heart, or because of a shortage of oxygen in the body. The last one is Cardiomyopathy. It causes the heart's pumping ability to weaken; cardiomyopathy leads the cardiac muscles to become dysfunctional or alter their shape. This can lead to heart failure. In this paper, we have proposed an ensemble framework for heart disease prediction.

The outline of the rest of the paper is as follows: In the second Section, we have included the literature review in which we referred to various research works and explained the viability and performance of the different algorithms related to heart disease prediction. In the third Section we have explained different machine learning algorithms. In the fourth Section, the proposed framework has been explained in detail including model selection, parameter setting, Experimental setup & proposed methodology. In the fifth Section, performance metrics, comparison of the proposed framework with existing Machine Learning (ML) models & with existing literature is explained. Results are shown with respect to the existing model & literature. Sixth Section contains the conclusion and future scope.

## 2. Literature Review

In this section, we have taken 10 different research works and explained how other researchers have approached the problem and their different methodologies. The count of cardiovascular disease patients has been growing exponentially [1]. Researchers are using a range of approaches and algorithms to predict cardiovascular disease. Over the years, several types of research on the prognosis of cardiovascular illness have been carried out. Related studies are discussed below. In [8], authors have applied algorithms such as Logistic Regression, K-Nearest Neighbour, Naïve Bayes, Support Vector Machine (SVM), Artificial Neural network, Decision Trees, Random Forest & multi-layer perceptron (MLP). The authors have applied the dataset firstly on an individual basis. Afterwards, the dataset is applied together to form a clear picture.

Table 1: Comparison of existing approaches for heart disease prediction

| S. No. | Author (s) | Approach | Dataset | Accuracy |
|---|---|---|---|---|
| 1 | Modak et al.(2022) [14] | Multilayer perceptron | Cleveland, Hungarian, Switzerland, Long Beach, and Statlog | 87.70% |
| 2 | Sarah et al.(2022)[15] | Logistic regression | Cleveland | 85.25% |

| 3 | Nguyen et al. (2021) [16] | Naive Bayes, Logistic Regression, SVM and Decision Trees | Cleveland | 83.5% |
|---|---|---|---|---|
| 4 | Latha et al. (2019)[17] | Random Forest, Multilayer Perceptron, Bayes Net Naïve Bayes | Cleveland | 84.49% |
| 5 | Atallah et al. (2019) [11] | (SGD)Classifier, K-Nearest Neighbor Classifier, Random Forest Classifier, Logistic Regression Classifier | Cleveland | 90% |
| 6 | Pawlovsky (2018) [18] | Weighted k-nearest neighbour | Cleveland | 84.83% |
| 7 | Bialy et al. (2016) [19] | Ensemble of FDT,C4.5, MLP, SVM, and Naive Bayes | Cleveland | 85.30% |
| 8 | Miao et al. (2016) [20] | Adaptive boosting | UCI Repository | 80.14% |
| 9 | Bashir et al. (2014) [21] | Memory-based learner, DT-IG,DT-GI, Ensemble of Naive Bayes, and SVM | UCI Repository, ricco database | 88.52% |
| 10 | Detrano et al. (1989) [16] | Logistic regression-based discriminant function | Cleveland | 77.00% |

They have considered a wide range of factors of cardiovascular disease and used classification matrices to reach a robust solution. In [9], the authors have applied various algorithms and used K-fold (10-fold) cross-validation techniques on full and selected features to propose the model to build an Intelligent Hybrid Framework. For feature selection, the authors have used 4 distinct feature selection algorithms which include the LASSO feature selection technique, Relief feature selection technique, etc. After applying the feature selection techniques, the authors have calculated Chi-Square and P-value. Authors in [10] have used HGBDTLR, which is an algorithm based upon the stack. The authors have implemented the stack on Support Vector Machines, Decision Tree, Logistic Regression, Adaboost, Random Forest, Gradient Boosting Decision Tree, K-nearest Neighbour, and Hybrid Random Forest with Linear Model. The authors have evaluated indicators including Accuracy, Precision, Recall Score & F1 Score. The majority Voting Ensemble technique is used in [11], Several algorithms were used and the highest accuracy of 90% was reached. The authors have taken a majority vote of all the algorithms to improve overall accuracy. The authors have also calculated correlation with Target i.e. if a data point shows positive in the target how the other attribute relates to it. CHAID (Chi-Squared Automated Interaction Detector) a Decision Tree-like structure method is performed for multilevel splits of data in [12]. The CHAID decision tree decodes a patient's health status and provides cardiologists with a better understanding and capacity to discriminate between diseases. CHAID analysis of the dataset was performed to establish a better understanding. Afterwards, Majority voting was done to improve overall accuracy. Naïve Bayes, Logistic regression & Multilayer perceptron were used for voting. A summary of related studies is given in Table 1. Most of the work conducted by authors is by using machine learning algorithms to predict heart disease algorithms. Many attributes were considered by the authors but we have seen that the scale of the dataset is small in the majority of the cases. In this study, we have studied a variety of algorithms and techniques and applied them for cardiovascular disease prediction. In our work, we implement our algorithms on the dataset [13] of 1190 instances with 11 attributes and we show how algorithms perform at scale. Important attributes required for the experiment have been considered below.

**2.1. Our Contribution:**
- The proposed Framework consists of Stacking Based Ensemble learning which adds diversity to the classifier.
- A large dataset is used for training the ensemble model which helps in generalising the trained model.
- Hyper Parameter Tuning is used to select the best parameter for ML model training.
- The performance of the proposed framework is compared with existing literature on the basis of accuracy, precision, sensitivity, precision, F1 Score, ROC & MCC.

**3. Background & Preliminaries**

In this section, various machine learning classification methods that are used for the proposed framework are explained. Various Classifier models were tried before the final Ensembling of top performing models. 10 different classifiers were trained on the training data set. After the initial training 4 models were selected based on their accuracy measure.

A. Random Forest [22]

Random forest is a tree-based categorization algorithm. This method creates a forest with a large number of trees, as the name implies. It's an ensemble algorithm with many algorithms. A random sampling of the training set is utilized to generate a collection of decision trees. Before making a final choice based on majority voting, it repeats the operation with a range of random samples. When coping with missing values, the Random Forest algorithm is useful, although it can be overfitting. Proper parameter adjustment may be utilized to avoid overfitting.

B. Multi-layer Perception Classifier [23]

It is a type of artificial neural network (ANN). The multilayer perceptron (MLP) is a computer model modelled after the organic nervous system. It's employed in classification and regression issues where the dependent and independent variables have nonlinear relationships. A conventional neural network is made up of three layers: an input & an output layer and one or more hidden layer(s). Neurons are artificial nodes found in each layer (processing elements). Neurons are linked together, and by-passing data forward through the network, they compute values based on the inputs.

C. K-Nearest Neighbor (KNN) [24]

The method in the classifier finds the distances between the new exemplar and all of the training exemplars and then selects the nearest K data points to the new instance from a specified K number. Finally, the majority class of the K data points picked is used to classify the data. In this project, the K number was chosen as 9 because it delivered the best outcomes based on the GridsearchCV.

D. XGBoost [25]

XGBoost is a highly scalable gradient boosting system aimed at speed and performance. It differs from conventional gradient boosting algorithms in that it includes intelligent tree penalization, proportionate leaf node reduction, and additional randomization settings.

E. Support Vector Classifier [26]

For both regression and classification tasks, the Support Vector Classifier (SVC) or Support Vector Machine (SVM) is a frequently used supervised machine learning algorithm. It categorizes the two groups by determining the hyper-plane that best separates them. Support Vectors are individual observation coordinates. The Support Vector Machine (SVM) is a frontier that best divides the two classes by hyperplane. The goal of the SVM classifier is to maximize the margin. H1 does not separate the classes in the diagram below. H2 has a slight advantage, but only by a short margin. H3 separates them by the widest possible margin. The margin measures the distance between the separating hyper-plane and the training samples nearest to the hyper-plane.

F. Stochastic Gradient Descent [27]

The SGD technique was used to create a binary classifier. To obtain, the SGD technique takes random cases from the training set and computes the gradient based on that single occurrence, which is the minimum value of the cost function. Then, using the parameters selected to optimize the cost function, classification is performed using a simple binary classifier that can determine whether or not the cardiac disease is present.

G. AdaBoostClassifier [28]

It is a meta-assessor that starts by fitting a classifier on the first dataset, then, at that point, fits extra duplicates of the classifier on the equivalent dataset while changing loads of ineffectively arranged examples so that succeeding classifiers centre more around troublesome cases.

H. Classification & Regression Trees (CART) [29]

The phrase CART analysis refers to both classification and regression tree analysis. Classification tree analysis is performed when the intended outcome is the class (discrete) to which the data belongs. Regression tree analysis is used when the expected outcome is a real number. These are the several kinds of decision trees. The goal is to learn basic decision rules from data attributes and build a model that predicts the value of a target variable. A tree is a piecewise constant approximation.

I. Gradient Boosting Machines [30]

The GBM is a numerical optimization method for identifying an additive model that minimizes the loss function. As a result, the GBM technique builds a new decision tree that optimally decreases the loss function at each step. In regression, the process begins with the model being initialized with a first estimate, which is often a decision tree that minimizes the loss function, and then a new decision tree is fitted to the current residual and added to the prior model to update the residual at each step. Until the user specifies a maximum number of iterations, the algorithm continues to iterate. This is a step-by-step method, which means that the decision trees that were used to build the model in earlier steps are not modified in later steps. The model is improved in areas where it does not perform well by fitting decision trees to the residuals.

J. Naive Bayes Classifiers [31]

The administered learning algorithms based on Bayes' technique with the "naive" supposition of conditional independence between each pair of features given the value of the class variable are known as naive Bayes methods. When compared to more complex algorithms, Naive Bayes classifiers can be exceedingly fast. Because the class conditional feature distributions are decoupled, each distribution may be estimated as a one-dimensional distribution independently. This, in turn, aids in the alleviation of problems caused by the high dimensionality.

## 4. Proposed Framework

In this section, we explained model selection criteria and parameter setting of different algorithms used in building the framework. The experimental setup & proposed methodology has been explained further.

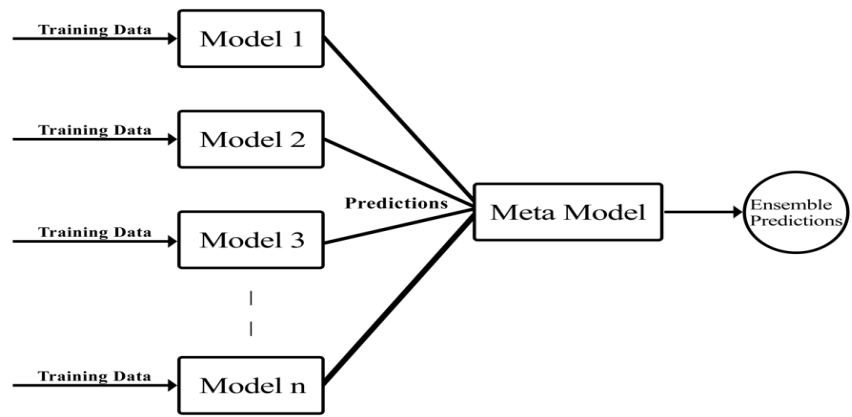

Figure 1: The proposed ensemble framework for heart disease prediction

### 4.1. Model Selection

Figure 1 comprehensively shows the proposed framework for heart disease prediction. Firstly, we have taken Data from IEEE Data Port [13] which is explained in Section 3.3. After deeply analysing the data by finding correlations among other attributes, we have removed outliers. Then split our data into two partitions: Training Data consisting of 80% of instances &Testing Data with 20% instances.

After we build our Model, we performed stacking also known as Stacked Regression which is a class of algorithms that involves training a second-level meta learner to find the optimal combination of base learners. Stacking is different from bagging and boosting, the goal of stacking is to ensemble strong, diverse sets of learners together. We have used numerous baseline models by using cross-validation with 10-folds to choose the best models among others. After which we input our pre-processed data into the base learners to generate ensemble predictions. As per the performance of different baseline models on cross-validation accuracy, we select the best performing models for the stacked ensemble so that their ensemble will produce higher performance in comparison to individual machine learning models. Finally, we compare our results with the base learners and various existing recent approaches for heart disease prediction.

### 4.2. Parameter Setting

In this section, we have explained diverse parameters used to boost accuracy across our Stacked Ensemble model. In XGBoost, we have tried different values of n estimators which are 100, 500, 1000, and 2000 out of which we received the best accuracy from the value of n estimator as 500. In KNN, we ran our model with different values of K and we received the best accuracy for k=9. In Random Forest Algorithm, Out of Gini and Entropy, we have used Entropy criteria to improve accuracy. In Extra Tree Classifier we have tried different values of n estimators which are 100, 500 & 1000 out of which we received the best accuracy from the value of n estimator as 500.

### 4.3. Experimental Setup

#### 4.3.1 Data set

The IEEE Data Port was used to retrieve the cardiovascular disease (CVD) dataset [13] [32]. The dataset was compiled by merging five prominent CVD datasets that were previously available individually named Hungarian, Cleveland, Long Beach VA, Switzerland & Statlog (Heart) Dataset but never combined before as far as we know. The dataset comprises 11 features of 1190 instances. The attributes in dataset, Age (in years), Level of Blood Pressure (in mm Hg), Serum Cholesterol (in mg/dl), Maximum Heart Rate Achieved (71–202), Old Peak =ST (depression) are of numerical data types and the attributes Sex (0,1), Chest Pain Type (1,2,3,4), Fasting Blood Sugar (1,0 > 120 mg/dl), Resting Electrocardiogram Results (0,1,2), Exercise-Induced Angina (0,1), The Slope of the Peak Exercise ST-Segment (0,1,2), Target (0,1) are of nominal data types. The description of nominal attributes is shown in Table 2.

Table 2: Description of Nominal Attributes

| Attribute | Description |
| --- | --- |
| Sex | Gender of patients (1 = male, 0= female) |
| Resting electrocardiogram results | Results of ECG while at rest<br>-- Value 0: normal |

|  | -- Value 1: having ST-T wave abnormality (T wave inversions and/or ST elevation or depression of > 0.05 mV)<br>-- Value 2: showing probable or definite left ventricular hypertrophy by Estes' criteria |
|---|---|
| Chest Pain Type | Types of chest pain experienced by patients categorised:-<br>-- Value 1: typical angina<br>-- Value 2: atypical angina<br>-- Value 3: non-anginal pain<br>-- Value 4: asymptomatic |
| The Slope of the Peak Exercise ST Segment | ST segment measured in terms of slope during peak exercise<br>-- Value 1: upsloping<br>-- Value 2: flat<br>-- Value 3: down sloping |
| Exercise induced angina | Angina induced by exercise (1 = Yes; 0 = No) |
| Fasting Blood sugar | (Blood sugar levels on fasting > 120 mg/dl) (1 = true; 0 = false) |
| class | 1 = heart disease, 0 = Normal |

To examine the dataset, relationship esteem was determined between Target Diagnosis and each of the values. It tends to be noticed that the most correlated features with the target attributes were ST Slope, Exercise-Induced Angina, and Chest Pain Type & ST Depression as shown in Table 3. Figure 2 shows the heat map displaying the correlation between all characteristics is displayed to provide a clear picture of the feature connection between every one of the characteristics. In addition, Figure 3 shows a pie chart depicting the gender distribution of the cases in the Heart Disease Dataset (Comprehensive). The statistics clearly show that males outnumber females by 76% to 24%. Furthermore, to preview the data given in the continuous attribute visualization, histograms are plotted from Figures 4 to11. In addition, Figure 12 shows, that the highest correlated continuous property (ST Slope) is plotted against age to see if there is any correlation. People with an ST slope of 2 are more likely to get heart disease, regardless of their age.

### 4.3.2 Data visualisation & Correlation of data attributes

Table 3: Correlation with Target Diagnosis

| | |
|---|---|
| st_slope | 0.505608 |
| exercise_induced_angina | 0.481467 |
| chest_pain_type | 0.460127 |
| st_depression | 0.398385 |
| sex | 0.311267 |
| age | 0.262029 |
| fasting_blood_sugar | 0.216695 |
| resting_blood_pressure | 0.121415 |
| rest_ecg | 0.073059 |
| cholesterol | -0.198366 |
| max_heart_rate_achieved | -0.413278 |

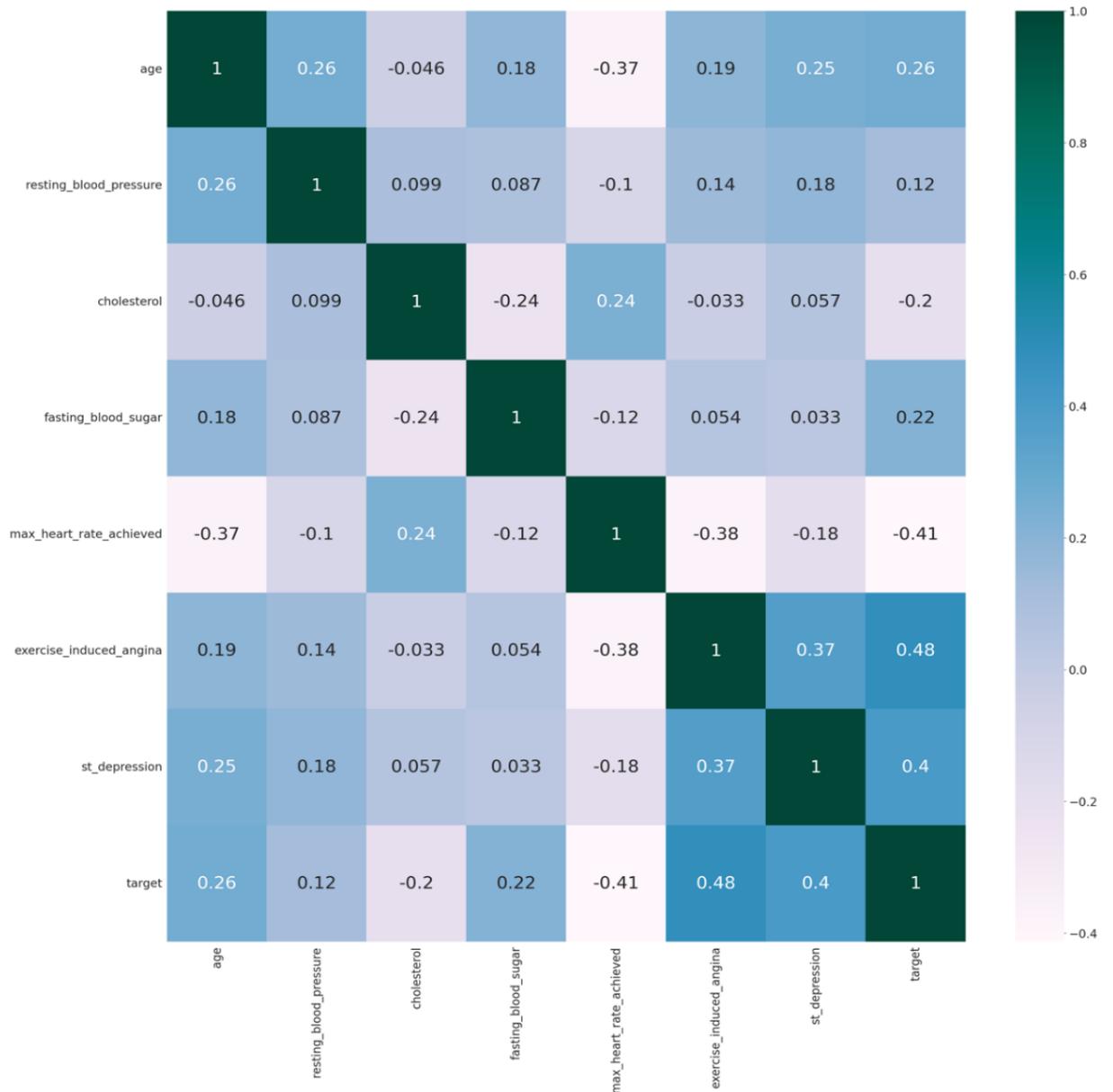

Figure 2: Cross-correlation values through Heat map

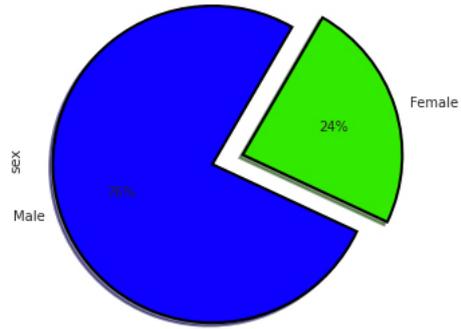

Figure 3: Gender distribution within the dataset

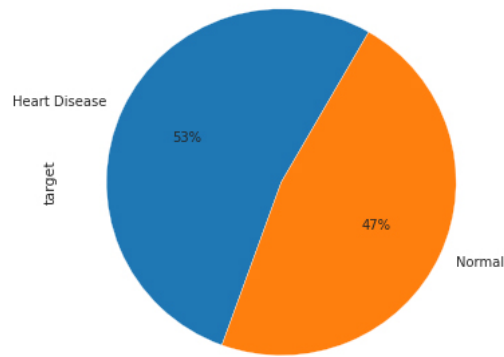

Figure 4: Number of Heart Disease Patient in the dataset

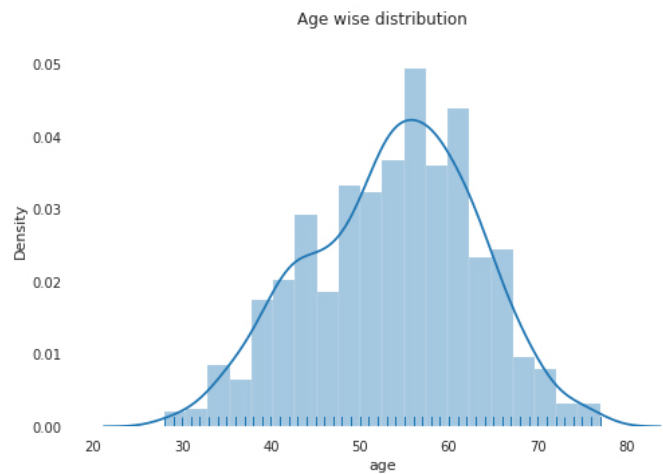

Figure 5: Age wise distribution of the Dataset

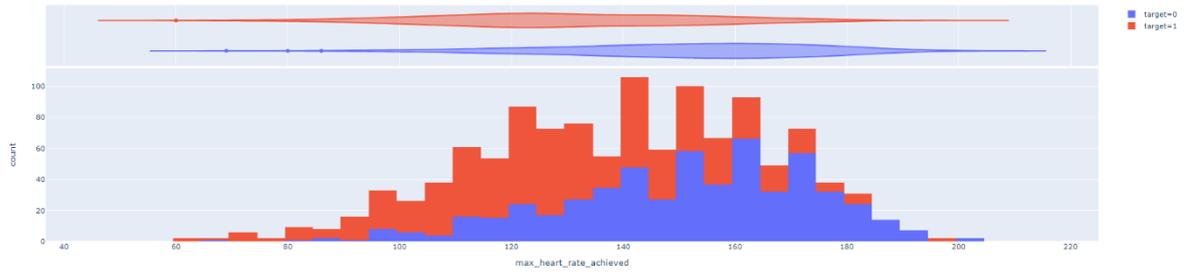

Figure 6: Distribution of Maximum Heart Rate Achieved.

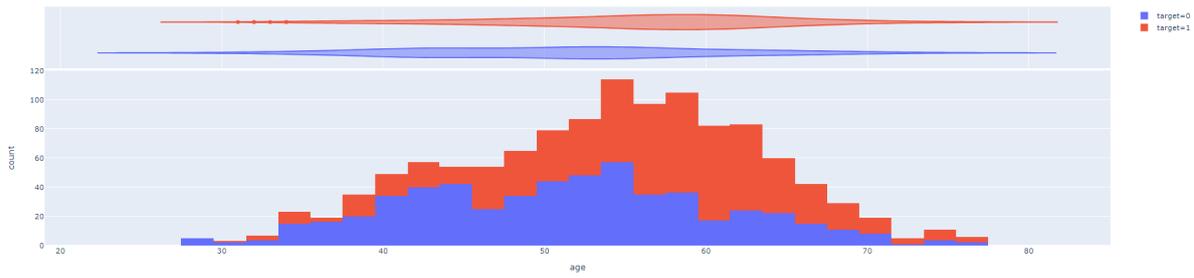

Figure 7: Distribution of Age vs Target

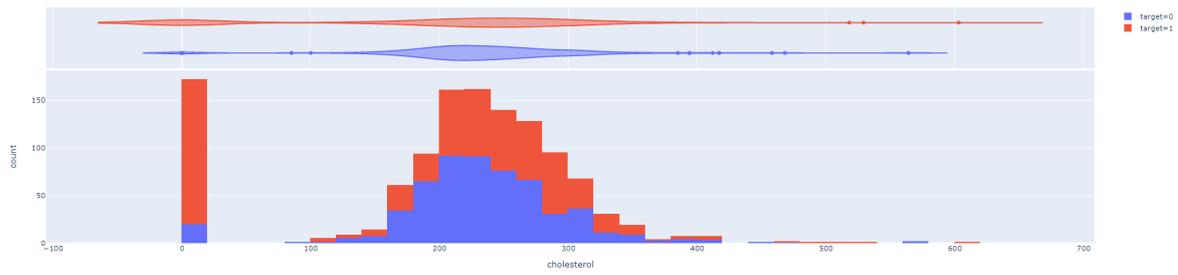

Figure 8: Distribution of Cholesterol.

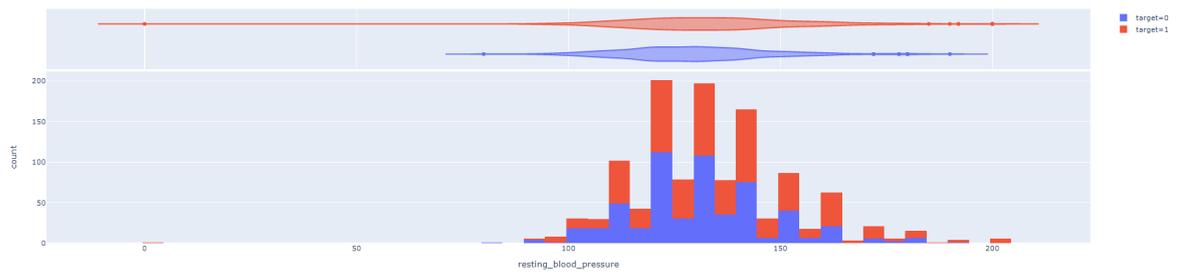

Figure 9: Distribution of Resting Blood Pressure.

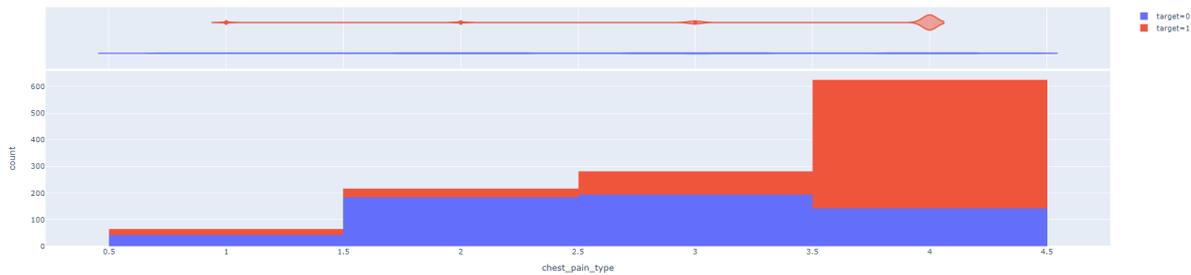

Figure 10: Distribution of Chest Pain.

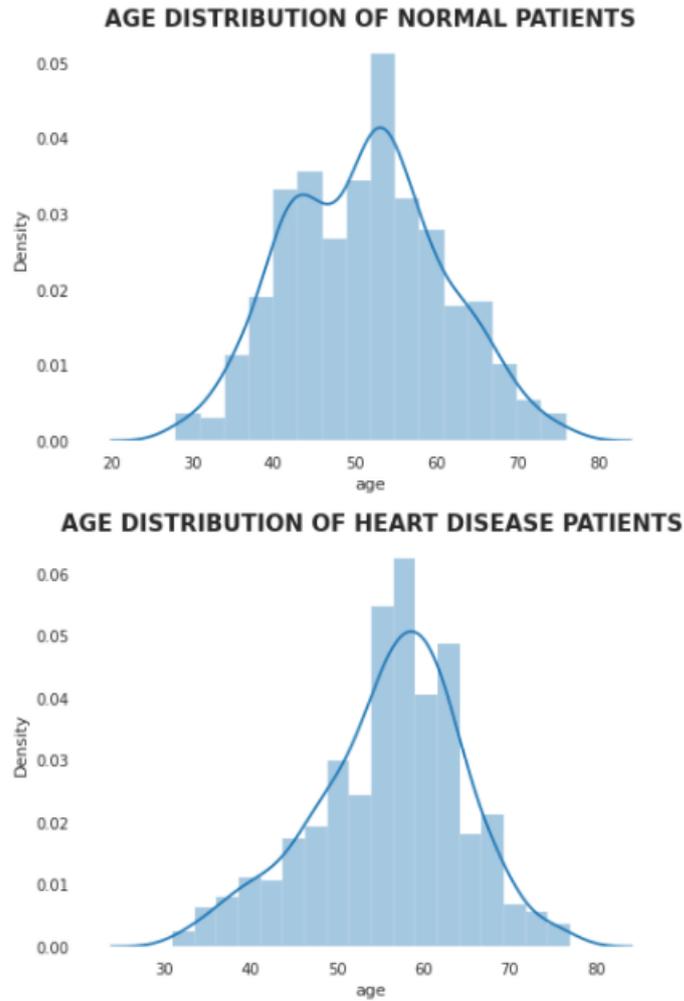

Figure 11: Age wise distribution of Normal Patients & Heart Disease patients
.

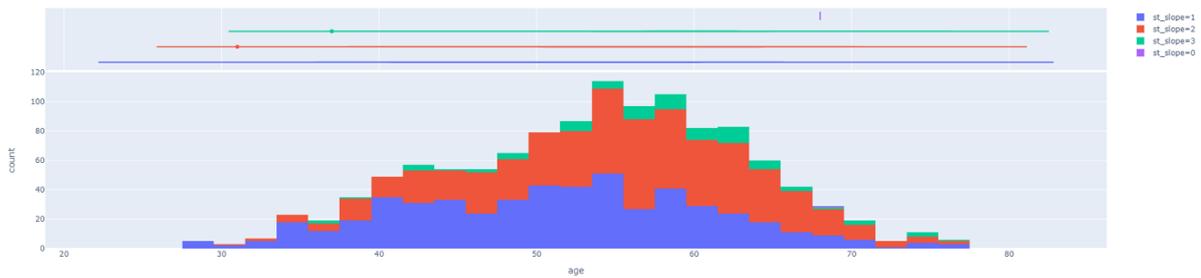

Figure 12: ST Slope plotted against age

### 4.3.3 Proposed Methodology

Stacking, also known as Super Learning is an ensemble technique that involves training a "meta learner" through a combination of multiple classification models. the purpose of stacking is to ensemble strong, varied sets of learners together. The algorithm for stacking is shown in Algorithm 1 and variables are defined in Table 4.

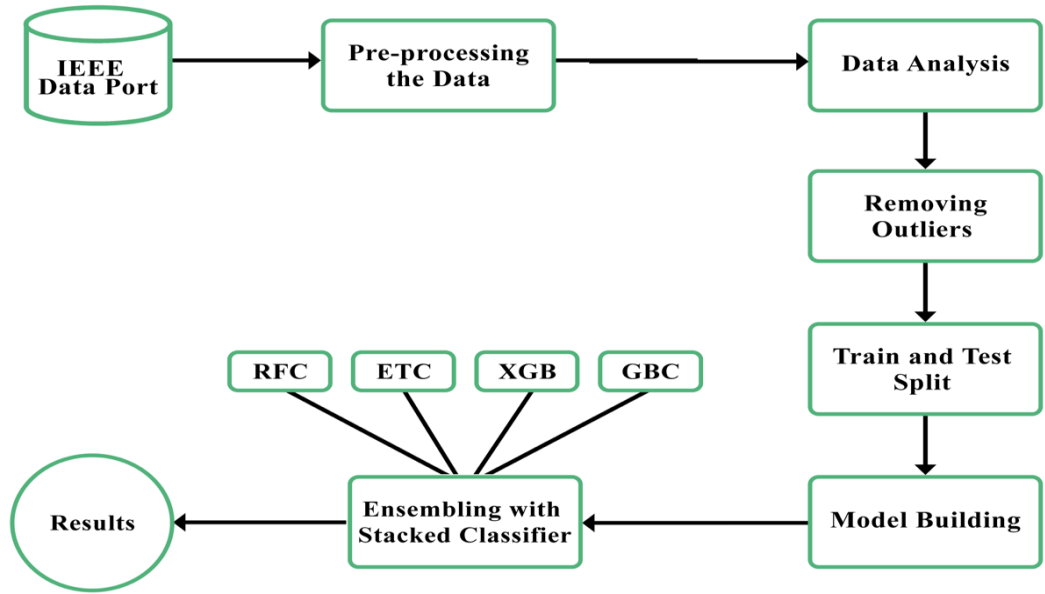

Figure 13: The architecture of proposed ensemble model

**Algorithm 1:** Proposed Ensemble framework for heart disease prediction

```
Let H = {h₁, h₂, h₃ … hₙ} be the given dataset
E = {E₁, E₂, E₃, ... Eₙ}, the set of Machine learning ensemble classifiers
X = the 80% dataset for training, X ∈ H
Y = the 20% dataset for testing, Y ∈ H
Z = meta level classifier
D = n(H)
for j=1 to D do
Begin
    M(j) = E(j) used for training the Model on X
    Next j
    M = M ∪ Z
End
Result = M classifies Y
```

Table 4: Symbols used in Algorithm 1

| S. No. | Symbols | Meaning |
|---|---|---|
| 1. | h | Attributes of dataset |
| 2. | E | Machine learning classifiers |
| 3. | X | Training set |
| 4. | Y | Testing set |

| 5. | Z | Meta level classifier |
| 6. | D | Number of attributes in dataset |
| 7. | j | Iterator Variable |
| 8. | M | Model |

In the proposed ensemble model, firstly, the original data is fed into several different models. The Meta classifier is then utilized to estimate the input as well as the output of each model, and also the weights are estimated. The top-performing models are chosen, while the rest are rejected. By utilizing a meta classifier, Stacking is the combination of numerous base classifiers learned on a single dataset using various learning methods. To make ensemble predictions, train the base learners and input the predictions into the meta learner, as shown in Figure 1. Figure 13 comprehensively shows our work i.e. first, we have taken Data from IEEE Data Port which is shown in Data Collection & Pre-processing after deeply analyzing the data and finding correlations among other attributes we have removed outliers. Then spitted our data in 80% & 20% values.

## 5. Results & Discussion

In this section, we have discussed the results & analysis of our proposed framework. Different performance metrics have been used to evaluate the algorithms. Further, we have compared our model with other existing models and its comparison with respect to the accuracy, precision, sensitivity, precision, F1 Score, ROC & MCC. We have also discussed the proposed model with different algorithms and models covered in Section 2.

### 5.1 Performance Metrics

True positives (TPs), true negatives (TNs), false positives (FPs), and false negatives (FNs) are the performance measures mentioned here, as stated below:

Sensitivity or True positive rate (TPR): It describes the possibility of a classifier correctly predicting a positive result when illness is present. It's calculated as follows:

$$Sensitivity = \frac{TP}{TP + FN}$$

Specificity or True negative rate (TNR): When there is no illness, it is a classifier's probability of predicting a negative outcome. It's calculated as follows:

$$Specificity = \frac{TN}{TN + FP}$$

Accuracy: It is one of the most commonly used measures used to evaluate a classifier's performance. It's calculated as a percentage of correctly identified samples, and it's written as:

$$Accuracy = \frac{TP + TN}{TP + TN + FP + FN}$$

Mathews correlation coefficient (MCC): It is a correlation coefficient between observed and predicted classes. MCC 1⁄4 +1 indicates a flawless prediction, MCC 1⁄4 0 indicates no better than the random prediction, and MCC 1⁄4 1 indicates complete disagreement between observed and projected values. Even though the class sizes are substantially different, this statistic is generally considered a balanced measure. MCC is defined as follows:

$$MCC = \frac{TP.TN - FP.FN}{\sqrt{(TP+FP).(TP+FN).(TN.+FP).(TN+FN)}}$$

AU-ROC (area under the receiver operating characteristic curve): For classification difficulties, it is also a useful and widely used performance metric. It is plotted using TPR vs FPR at various threshold values. Because it evaluates performance across a wide variety of class distributions and error levels, the AU-ROC is an excellent metric for performance comparison. It is defined as follows:

$$AU - ROC = 1/2(\frac{TP}{TP+FN} + \frac{TN}{TN+FP})$$

F1 score: It's viewed as the precision and recall's weighted average (or harmonic mean). An F1 score of 1 is regarded as the greatest, while a score of 0 is considered the worst. The TNs are not taken into consideration in F-measures. The F1 score may be calculated as follows:

$$F1\ score = 2\ X\frac{Precision X Recall}{Precision + Recall}$$

## 5.2 Comparison with ML Models

We built numerous baseline models and used cross-validation with 10-folds to choose the best models. Models with high accuracies are used in the stacking approach. The accuracy of the baseline models is shown in Table 5 and their graph is plotted as shown in Figure 14. As we have observed from Table 5 and its graph, the best performing algorithms are Random Forest, Gradient Boosting Classifier, XGBoost & Extra Tree Classifier. These algorithms are stacked as shown in Figure 14. As shown in Table 6, we have tested the algorithms based on accuracy. Apart from that precision, sensitivity, specificity, F-1 Score & Mathews correlation coefficient (MCC) are also measured for evaluating performance. From Table 6, the stacked classifier has a greater classification accuracy of 92.34 per cent than the other classifiers. Considering other factors, the stacked classifier has the highest F1 Score, MCC with values of 92.74% and 84.64% respectively. As shown in Figure 15, the stacked classifier has the highest specificity of 91.07% and second highest Sensitivity of 93.49% whereas the random forest has the highest Sensitivity of 95.12%. Again, stacked classifiers have the highest Precision and ROC values of 92.00% and 92.28% respectively as shown in Figure 16.

Additionally, to analyse the models developed, a ROC (receiver operating characteristic curve) was constructed for each model, as shown in Figure 17. The ROC shows the classifier's diagnostic abilities, Figure 17 shows the area under each curve computed and presented. The better the model's diagnostic capacity, the closer the area value of the ROC curve is to one. The precision-recall curve depicts the trade-off between precision and recall for different thresholds. The average precision recall of the 4 models is shown in figure 18. A low false-positive rate is associated with good accuracy, and a low false-negative rate is highly associated with recall.

Table 5: Comparison of ML algorithm & their respective accuracies

| S. No. | Algorithm | Accuracy |
|---|---|---|
| **1** | **XGB** | **91.91%** |
| **2** | **ExtraTree Classifier** | **90.93%** |
| **3** | **Random Forest** | **90.21%** |
| **4** | **GBM** | **84.25%** |
| 5 | CART | 84.25% |
| 6 | MLP | 84.25% |
| 7 | Adaboost | 83.40% |
| 8 | SVC | 82.55% |
| 9 | SGD | 82.12% |
| 10 | KNN | 80.85% |

Figure 14: Comparison of accuracy of proposed framework with different ML models

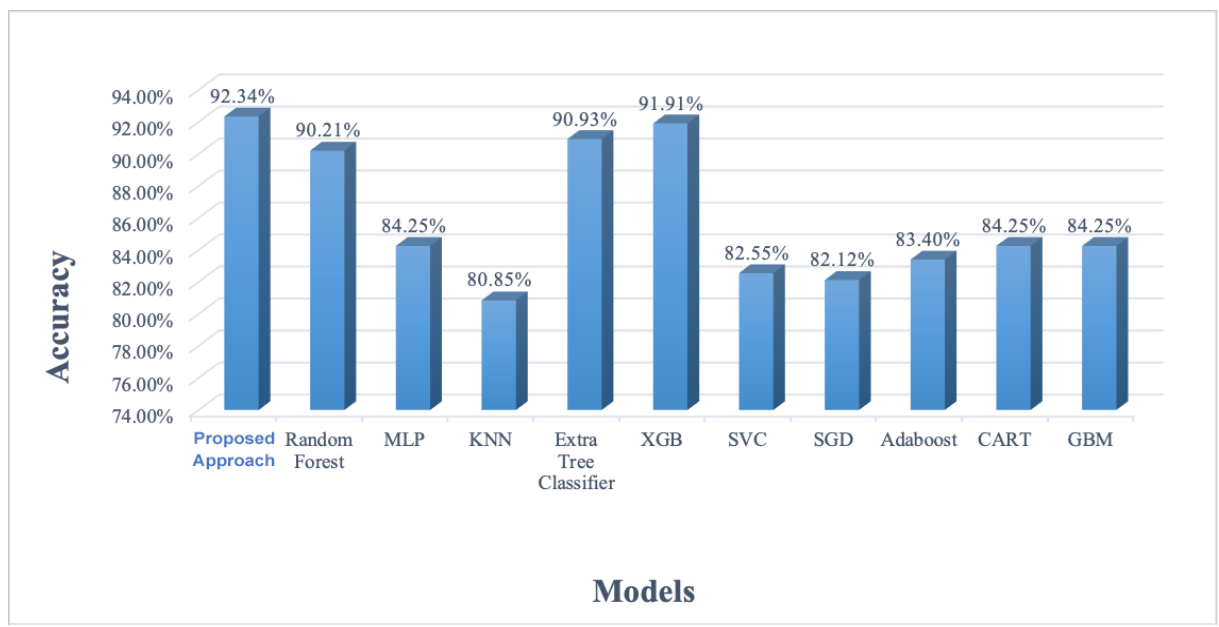

Table 6: Comparison of proposed framework with existing ML models

| Model | Accuracy | Precision | Sensitivity | Specificity | F1 Score | ROC | MCC |
|---|---|---|---|---|---|---|---|
| **Proposed Approach** | **92.34%** | **92.00%** | **93.49%** | **91.07%** | **92.74%** | **92.28%** | **84.64%** |

| Random Forest | 90.21% | 87.31% | 95.12% | 84.82% | 91.05% | 89.97% | 80.65% |
| MLP | 84.25% | 82.08% | 89.43% | 78.57% | 85.60% | 84.00% | 68.60% |
| KNN | 80.85% | 78.67% | 86.99% | 74.10% | 82.62% | 80.54% | 61.80% |
| Extra Tree Classifier | 90.93% | 88.54% | 94.30% | 86.60% | 91.33% | 90.45% | 81.36% |
| XGB | 91.91% | 90.62% | 94.30% | 89.28% | 92.43% | 91.79% | 83.83% |
| SVC | 82.55% | 80.14% | 88.61% | 75.89% | 84.16% | 82.25% | 65.25% |
| SGD | 82.12% | 80.00% | 87.80% | 75.89% | 83.72% | 81.84% | 64.34% |
| Adaboost | 83.40% | 81.43% | 88.61% | 77.67% | 84.82% | 83.14% | 66.88% |
| CART | 84.25% | 83.59% | 86.99% | 81.25% | 85.25% | 84.12% | 68.44% |
| GBM | 84.25% | 81.61% | 90.24% | 77.67% | 85.71% | 83.96% | 68.70% |

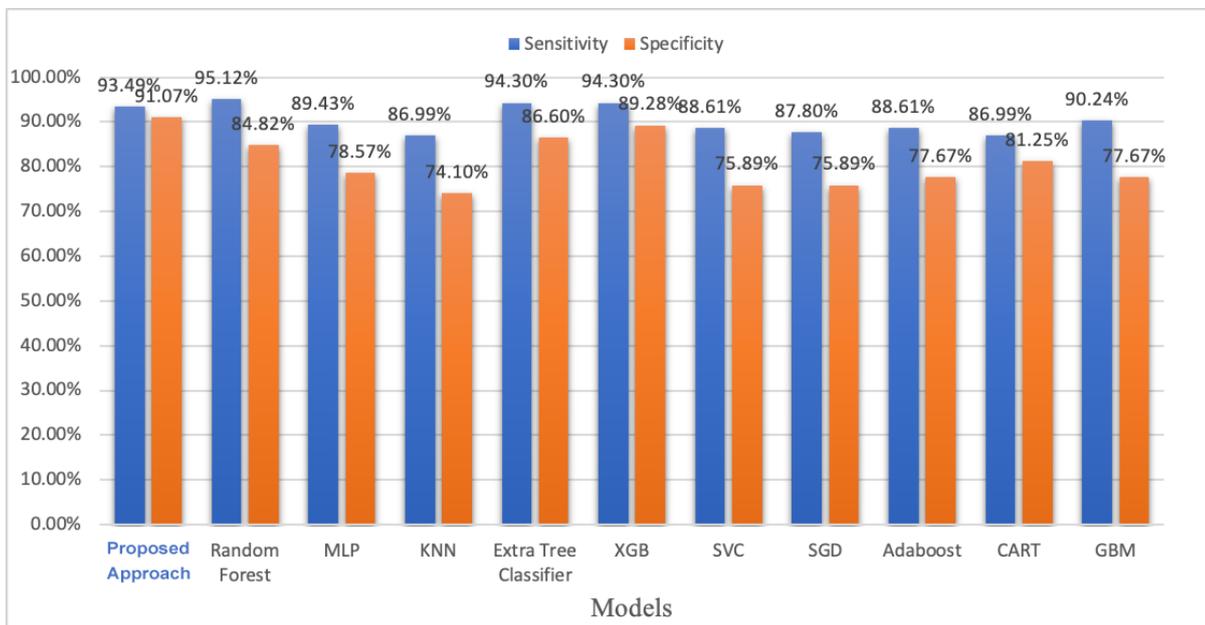

Figure 15: Comparison of sensitivity and specificity of proposed framework with different ML models

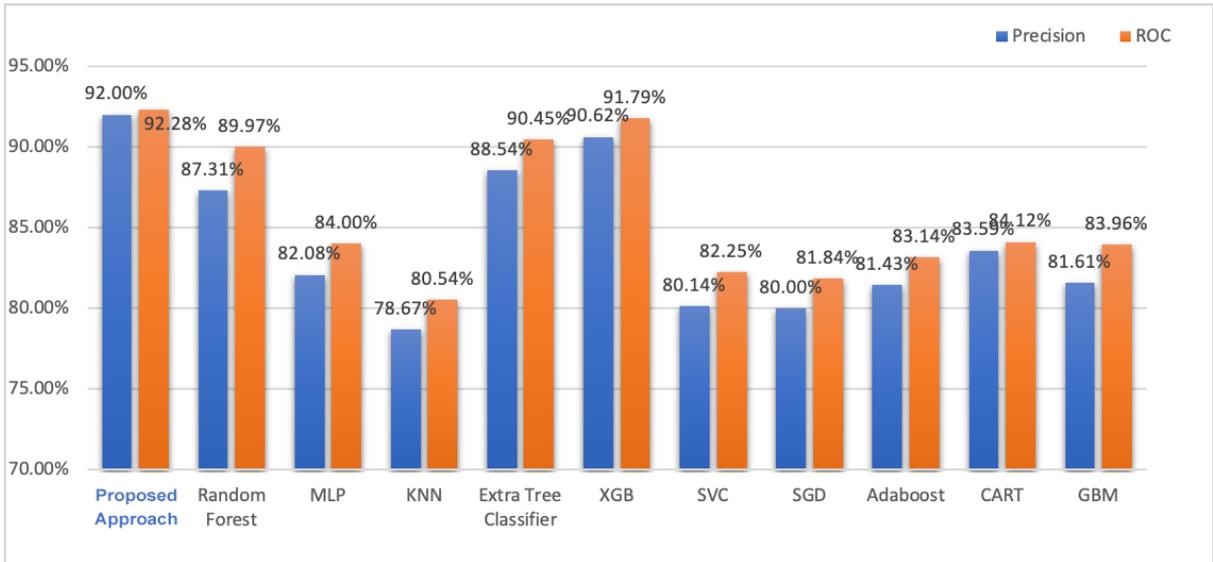

Figure 16: Comparison of precision & ROC of proposed framework with different ML models

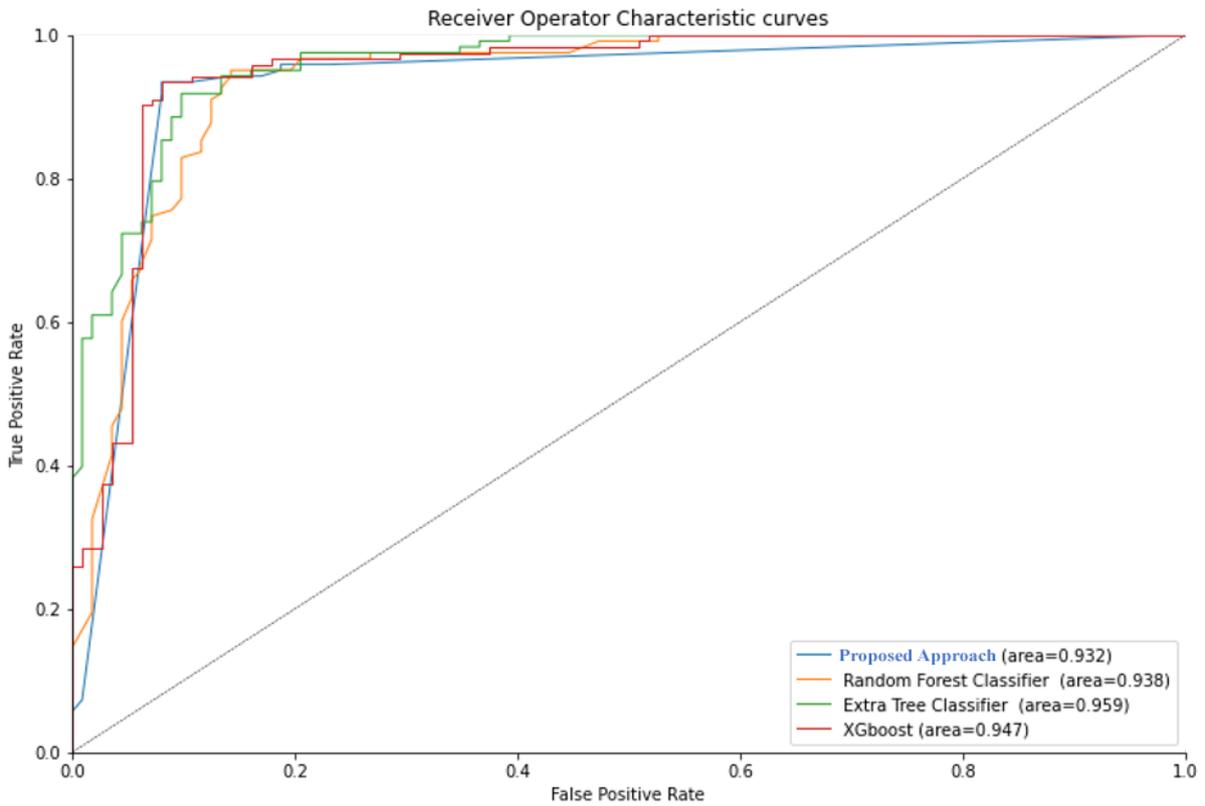

Figure 17: AUC-ROC curves for proposed framework and other classifiers

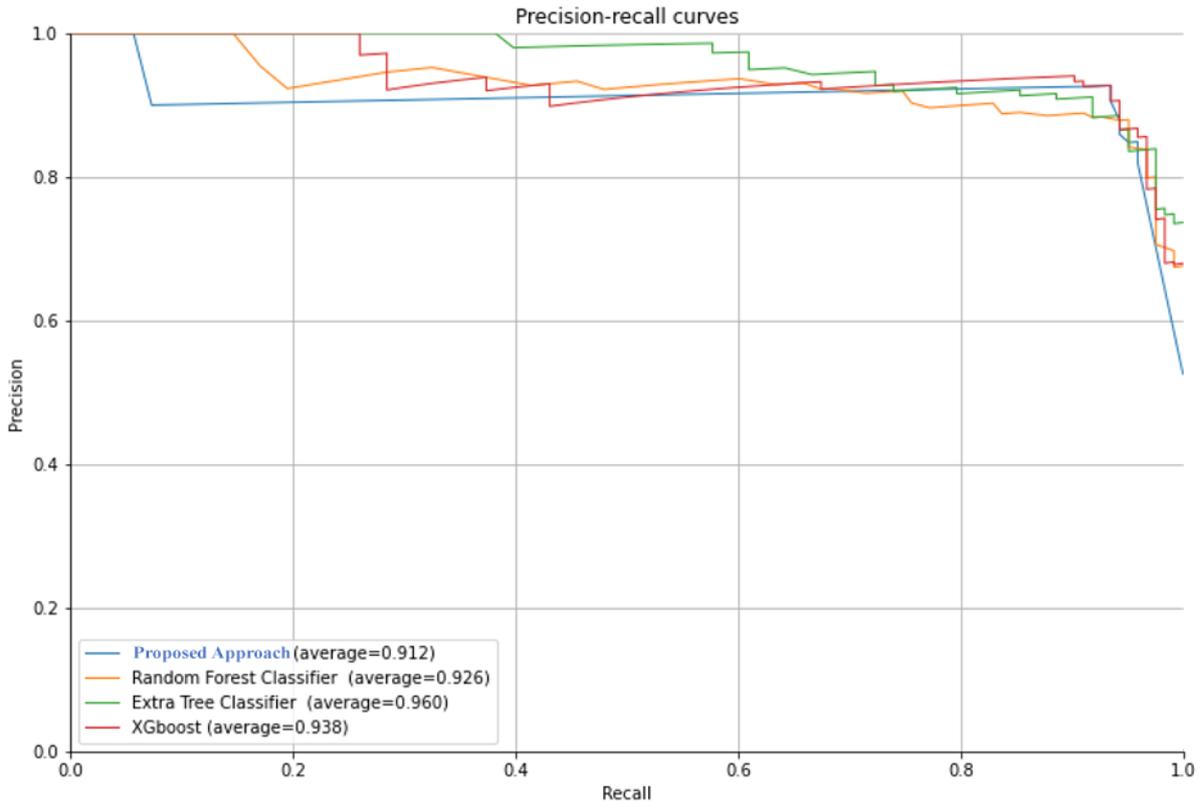

Figure 18: Precision-Recall curve for the heart disease predicting classifiers

## 5.3 Comparison with Existing Literature

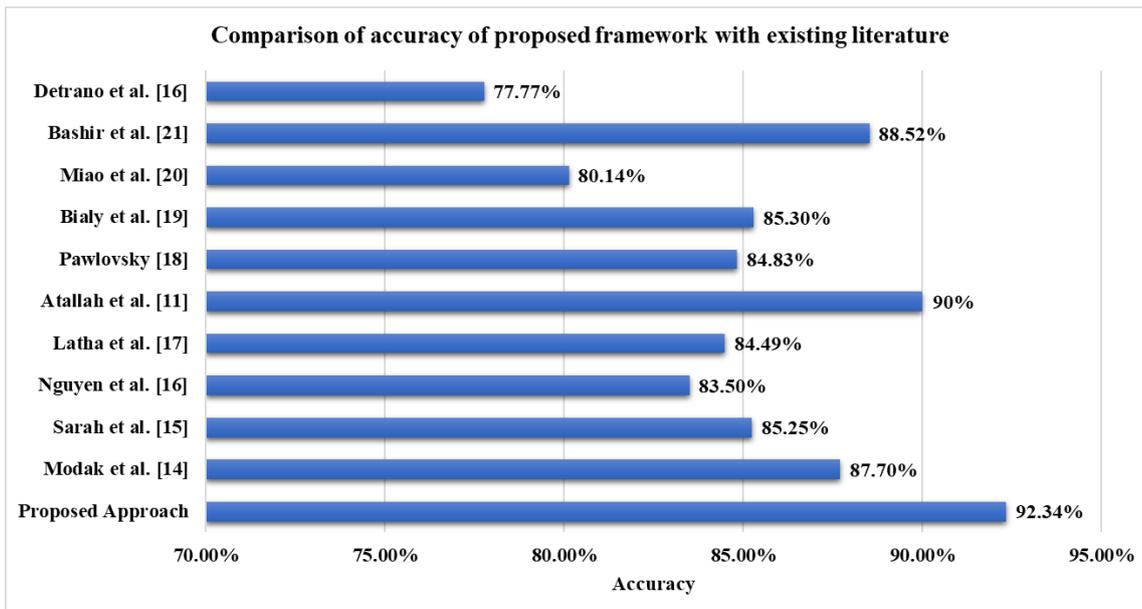

Figure 19: Comparison of results with others approaches covered in Table 1.

The comparison, in this case, shows an important part that can help medical research and diagnosis because of the relative nature of all the algorithms. Figure 19 completely shows the diverse nature and performances of all the algorithms. When we apply this stacked approach in real life, the medical community can get the benefit from this diversity by knowing how an algorithm performs in different cardiovascular diseases.

## 6. Conclusion and Future Work

In countries like India, where resources are limited and population numbers are rising. The need for better health care is alarming, as we have seen in the case of Covid-19. The proposed framework can help the early diagnosis of the patients and can equally contribute to the healthcare domain to early predict cardiovascular disease in this alarming situation. The epistemic evidence & results of our proposed framework are robust in nature because of its stacked machine learning based approach. Our proposed framework outperforms the existing state-of-art literature as shown in the results section. In future, we would like to explore our approach to the large dataset using deep learning concepts.

**Acknowledgement**

The authors are grateful to Mr. Saransh Rohilla & Mr. Suveer Sharma for their assistance.